\documentclass{ncjms}
\usepackage{verbatim}

\begin{document}

% input the title of the manuscript
\title{A Survey on the Visual Perceptions of Gaussian Noise Filtering on Photography}
\titlerunning{Investigating Image Quality Loss $\ldots$}   %specify header with (shorter) title

% specify the author(s) of the manuscript
  \author[Aidan J. Draper]{Aidan J. Draper}
  \address[Aidan J. Draper]{Department of Math and Statistics, Elon University, Elon, NC 27244, US}
  \email[Corresponding author]{adraper2@elon.edu}
  \urladdr{http://www.aidandraper.com/} % Delete if not wanted.

  %second author
  \author{Laura L. Taylor}
  \address[Laura L. Taylor]{Department of Math and Statistics, Elon University, Elon, NC 27244, US}
  \email{ltaylor18@elon.edu}

\authorsrunning{A. J. ~Draper and L. L. Taylor}  %specify header with author names, use "et al." if too many

\begin{abstract}
Statisticians, as well as machine learning and computer vision experts, have been studying image reconstitution through denoising different domains of photography, such as textual documentation, tomographic, astronomical, and low-light photography. In this paper, we apply common inferential kernel filters in the R and python languages, as well as Adobe Lightroom's denoise filter, and compare their effectiveness in removing noise from JPEG images. We ran standard benchmark tests to evaluate each method's effectiveness for removing noise. In doing so, we also surveyed students at Elon University about their opinion of a single filtered photo from a collection of photos processed by the various filter methods. Many scientists believe that noise filters cause blurring and image quality loss so we analyzed whether or not people felt as though denoising causes any quality loss as compared to their noiseless images. Individuals assigned scores indicating the image quality of a denoised photo compared to its noiseless counterpart on a 1 to 10 scale. Survey scores are compared across filters to evaluate whether there were significant differences in image quality scores received. Benchmark scores were compared to the visual perception scores. Then, an analysis of covariance test was run to identify whether or not survey training scores explained any unplanned variation in visual scores assigned by students across the filter methods.
\end{abstract}

%
% Include AMS Subject Classifications as can be find on http://www.ams.org/msc/
 %separated by semicolon

%
% Include keywords for the paper, separated by semi-colons; ending with a point.

%\date{\today}

\maketitle

%

%
%%%%%%%%%%%%%%%%%%%%%%%%%%%%%%%%%%%%%%%%%%%%%%%%%%%%%%
%%%%%%%%%%%%%%%%%%%%%%%%%%%%%%%%%%%%%%%%%%%%%%%%%%%%%%
%
%  Actual body of the manuscript starts here
%
%%%%%%%%%%%%%%%%%%%%%%%%%%%%%%%%%%%%%%%%%%%%%%%%%%%%%%

\section{Introduction}
The world is full of many naturally-occurring signals. Because there are so many, it becomes rather hard to isolate them at times. Photographers try to isolate light signals during exposure, which can often be difficult. Many other domains of science (Ecology: \citealp{statespace16}; Microscopy: \citealp{perfeval16}; Tomography: \citealp{imgproctom06}; Astronomy: \citealp{1502901}; Physics: \citealp{939832}) study signals that can be even harder to single out. Even with the immense collection of advanced electronics found in today's average digital camera, captured light signals are still prone to random noise from sources such as the camera's sensor or circuitry and as a response to qualities of the environment, like temperature. This is what makes the field of signal denoising so compelling. In addition, no other physical environments resist noise as poorly as low-light environments do. In specific, dusk light is arguably the most detrimental to a photographer's image quality, yet it is still a popular time to shoot. This has increased the prevalence of noise in modern photography, which motivates this study. 

Many researchers have spent much of their life investigating methods for properly denoising signals. In doing so, standardized formulas have been developed to validate the effectiveness of these methods. However, past studies have reported that the benchmark scores typically associated with these methods are uncorrelated with unbiased observers' opinions of denoised-image quality \citep{blur, survey_blur, perf_eval}. These studies induced continued research into the investigation of image quality in this paper. 

In studying signal processing, it was apparent that there is a void in computer vision literature surrounding the visual perception of image quality in denoising methods. Additionally, this study differs from past work for three main reasons. First, it focuses on the subdomain of low-light photography due to current interests surrounding this setting in social media. This meant creating a unique low-light image dataset to suit the study. Second, quality of filtered grayscale images was investigated and visual perceptions of college students were collected. This was for convenience and to offer a different perspective than previous studies have had. Third, actual noise was captured rather than simulating the disruption of random pixels on a noiseless image. This adds some complications later, but also, provides real world examples of the filter methods' abilities. Lastly, a proprietary image denoising method from Adobe's creative cloud was included. This is probably the most accessible method to photographers and has also received a fair amount of criticism in the photography community, which made it an interesting addition to this study.

In this paper, five filter methods' performances on a single noisy photo were evaluated. After implementing these filtering methods on a single image, the benchmark scores of the filtered images were compared across methods. Then, a survey collected undergraduate college students' perceptions of the filtered images' quality in relation to a noiseless image. A one-way ANOVA test and an ANCOVA test were performed on the visual perception scores about the quality of filtered images in relation to their desired non-noisy state.
	
This paper is organized in the following way. This section proceeds with the background surrounding the subdomain of signal denoising in computer vision and, in doing so, the methods performed on the test image are shared. Section 2 describes the methodology and experimental design, which includes the R Shiny App survey that was built and the process in which the experiment was performed. Section 3 shares the results of the benchmark tests on the single image as well as survey participants perception of the filtered photos. A discussion about the experiment results is provided in Section 4. Finally, Section 5 discusses the conclusions that are drawn from the study.

\subsection{Background}
Computer vision is believed to be driven by two main pursuits for knowledge. From one perspective, scientists look to model human vision processes. Interest surrounds mimicking common human ability and understanding how human perception and comprehension occurs. On the other end of the spectrum, scientists look to improve autonomy in machines and perform advanced tasks, such as identifying objects or understanding dynamically-changing scenes, that is unrelated to understanding how human vision works \citep{zhang}. These philosophies inevitably overlap at times, but they are ultimately the motivators that drive the study of vision in computer science. The field is said to have emerged partly as a result of Larry Robert's thesis at MIT, where he introduced the concept of extracting 3-dimensional shapes from 2-dimensional images using ``line drawing" to retrieve edge information \citep{larry}. Many scientists would follow in his footsteps in studying a subdomain of computer vision that is today known as edge detection. Self-driving vehicles are some of the newest technology that require advanced methods for scene understanding that mimic human perception, while also performing many other processes that go beyond our human capability, like tracking distance from other vehicles. This paper analyzes the work of one sub-field of computer vision, image denoising.

The study of noise removal would be launched by the second motivator. Signal denoising researchers look to filter noise to improve pre-processing for machine autonomy and a broad scope of other science processes. Image denoising has been used to cleanup tomographic photos for electron identification \citep{fernandez}. It has also been used to process and cleanup many other signals, including electronic hisses, magnetized particles from magnetic film, tendrils and candlesticks from stock market data, and grain from satellite images. The specific study of digital image denoising emerged after the invention of the first charge-couple device camera in 1975. These charge-coupled devices would allow electronic storage of images and later, the computation of pixels. Charge-coupling devices can still be found in a wide variety of devices, including most modern compact lens and digital single-lens reflex cameras, computer vision robots, and satellites.

% need to add more on the individual distributions
In the field of signal processing, there are a few different types of noise that haunt images. What makes statistics useful is that most types of noise follow the same probability density functions of some common random variables. The most common found in low-light photography are Gaussian noise and Poisson noise, which follow the distribution of their respective random variables.

There are many heuristic methods for noise removal, but none have been able to fully denoise a scene to this date. Competition in the field emerged for the title of ``best-approach". Noise reduction is computationally demanding so methods are judged by their effectiveness and running-time. The need to evaluate algorithms with slightly-differing, approximate results led to the development of common scores for evaluating denoised photos. The most common include mean squared error (MSE), r-squared ($R^2$), peak signal-to-noise ratio (PSNR), structural similarity (SSIM), and run time. Their formulas are described below. 

\begin{equation}
    \text{MSE} = \frac{\mid \sum_{x} \sum_{y} \text{filtered state} - \sum_{x} \sum_{y} \text{true state} \mid^2} { \text{N}_{\text{True State}}}
\end{equation}

\begin{equation}
  R^2 = \frac{1 - \big( \sum_{x} \sum_{y}\text{true state} - \sum_{x} \sum_{y}\text{filtered state} \big)^2} {\big(\sum_{x} \sum_{y}\text{true state} - \mu_{\text{true state}} \big)^2}
\end{equation}

\begin{equation}  
  \text{PSNR} = 20* log_{10}\big(\frac{R^2}{\text{MSE}}\big)
\end{equation}  

\begin{equation}
   \text{SSIM} = \frac{\big( 2\mu_{\text{true state}}\mu_{\text{filtered state}} + c_1\big)*\big( 2\sigma_{\text{true state, filtered state}} + c_2\big)}{\big( \mu^2_{\text{true state}} + \mu^2_{\text{filtered state}} + c_1\big) * \big( \sigma^2_{\text{true state}} + \sigma^2_{\text{filtered state}} + c_2\big)}
\end{equation}    
\newline

\noindent{\textit{True state} and \textit{filtered state} describe the respective pixel matrices of the two images. Totals, means and standard deviations are evaluated based on the entirety of the image matrix. Higher PSNR values indicate better image restoration quality in most of cases. Lower MSE scores indicate less error between true image and filtered image, which is preferred. $R^2$ shares how much of the variation in the filtered image can be explained by the noiseless image, which is strongly influenced by what $(x,y)$ pixel values they share. Higher $R^2$ values indicate a stronger relationship between the noiseless image and the filtered image. Higher values of SSIM indicate a stronger structural simularity between the two images. SSIM evaluates photos similarly to PSNR and MSE, except it also considers the interdependence of pixels that are spatially similar.}

In this paper, the effectiveness of a few of the most common noise removal methods is explored, but for the purpose of cleaning up Gaussian noise, or `grain', from low-light images.

% I should further elaborate on each formula described
\subsection{Filter Methods}
There are many strategies implored in noise removal. Motwani et al. dissect and classify most methods currently used in the field into a tree graph \citep{Mukesh}. They classify every algorithm into two overarching categories: Spatial Domain and Transform Domain. Spatial domains implement box filters, while transform domains are slightly more complex in their classification, but can be generalized to relying on a standard basis function that differs from the traditional box filter. As mentioned earlier, the study was conducted using five of the most common, and arguably simplest, image denoising methods. They include the Three-by-three Mean Filter, Non-local Means Filter, Bilateral Filter, and two levels: 50\% and 100\%, of the Adobe Lightroom CC denoise filter. Each method will be dissected in this section.

\subsubsection{Three-by-three Mean Filter}
The Mean Filter is a rudimentary approach to noise filtering. It implements a box filter of size $z$ by $z$. This box filter is a smaller matrix that traverses the photo and calculates the mean for the center value of every three-by-three matrix that fits within the photo matrix. This box matrix is typically denoted as $x$ with a size of $n$, or $z*z$. The method is classified as a spatial domain linear filter known to be used to specifically target a decrease in mean squared error. The Mean Filter has received criticism for destroying edges, erasing fine details, and blurring lines in images \citep{Mukesh}. It is expressed symbolically as follows:

\begin{equation}
    I^{filtered}(x)=\frac{1}{n}\sum_{x_i\in \Omega}x_i
\end{equation}

\noindent {where $\Omega$ represents the entire image and $I^{filtered}$ is the resultant filtered image. This experiment implores the ability of a Three-by-three Mean Filter, which is a typical size for this method that causes less blurring than some of the larger spatial filter sizes.}

\subsubsection{Non-local Means Filter}
One of the first public introductions of the Non-local Means Filter was during an IEEE conference by \cite{baudes}. The algorithm relates to many linear filtering methods, but it calculates the weighted probability impact of each pixel in averaging the pixel of interest based on the similarity of neighboring pixel scores within the box instead of using a standard approach for probability weights, such as Gaussian probabilities or equal weights. It is expressed symbolically as:

\begin{equation}
    I^{filtered}(p)=\frac{1}{C(p)}\int_{\Omega}v(q)f(p,q)dq.
\end{equation}

\noindent{For this study, \citeauthor{opencv_library}'s \citeyearpar{opencv_library} OpenCV Non-local Means Filter was specifically implemented because their method has been optimized to decrease run time as much as possible.}

\subsubsection{Bilateral Filter}
The Bilateral Filter is a far more advanced filter that takes into account 3-dimensional space by also considering whether or not the next pixel would be unlikely to see based on a distribution of previously viewed pixels. This is expressed by the $g_s$ in the formula by calculating residual error. Symbolically, this formula is:

\begin{equation}
    I^{filtered}(x)=\frac{1}{W_p}\sum_{x_{i}\in \Omega}I(x_i)f_r(||I(x_i)-I(x)||)g_s(||x_i-x||)
\end{equation}

\noindent{Again, \citeauthor{opencv_library}'s \citeyearpar{opencv_library} OpenCV Bilateral Filter was specifically implemented rather than the traditional formula because OpenCV's method has been optimized for run time.}

\subsubsection{Adobe Lightroom Denoise Filter}
At the time of this study, there was no public information shared about Adobe's methods. It appears they have sacrificed some performance to be scalable on larger images and to output at an acceptable run time to users. In the author's opinion, at 100\%, their method appears to blur edges at an extreme rate. Photos appear washed and fine details are lost. 

\section{Methodology}
Noise is considered to be a quality that negatively impacts the perception of an image. One of the most typically seen random variable distributions of noise is Gaussian noise, which is an unwanted channel that is captured during the camera's acquisition of the desired light wavelengths. Many have explored methodology for replacing the unwanted pixels with computed pixels that replicate what the desired wavelengths may have shown using a variety of mathematical approaches for approximation and estimation. The common approach for testing these methods involves mimicking a noisy image and then, taking a true image with no noise so that they can compare how well their computed pixels replicate the photo that would be originally desired. They use the standard benchmark scores (PSNR, MSE, $R^2$, and SSIM) to score their results. 

This study models the process of analyzing methods' abilities to approximate missed pixels of the true signal, but goes a step further by surveying college undergraduates to investigate whether they believe image quality has in fact improved.

Introductory mathematics and statistics students attending Elon University were chosen as a sample population in order to expose younger students to undergraduate research. Additionally, it provided a convenient sampling frame for the time restrictions of this project. Under IRB approval, instructors from the previously mentioned courses were emailed and asked to share the survey link with students in their sections. Student participation was completely voluntary. Images of the survey are included \autoref{fig:survey} below. There were a total of nine slides in the survey, which are displayed in order in \autoref{fig:survey}. The survey included:
\begin{enumerate}
  \item Introduction slide - describes the survey, who to contact, and that response is optional,
  \item First instructions slide - informs users how to go about rating the following images,
  \item First true state image slide - a reference image to compare the following training images to,
  \item Training image one slide - an unfiltered noisy image with a horizontal score bar at the bottom,
  \item Training image two slide - a filtered noisy image with a horizontal score bar at the bottom,
  \item Training image three slide - the same true state image with a horizontal score bar at the bottom and a "Submit Part One" button that leads to the second part of the survey,
  \item Second instructions slide - informs the user how to go about rating the following image,
  \item Second true state image slide - a reference image to compare with the filtered image of interest,
  \item Filtered image of interest slide - a randomly-selected filtered image (either Three-by-three Mean, Non-local Means, Bilateral, Abobe 50\% or Adobe 100\%), a horizontal score bar to rate the image, and a "Submit Part Two" button to send the scores to a Google Sheet.
\end{enumerate}

\begin{figure}[ht]
  \centering
    \includegraphics[width=0.3\linewidth]{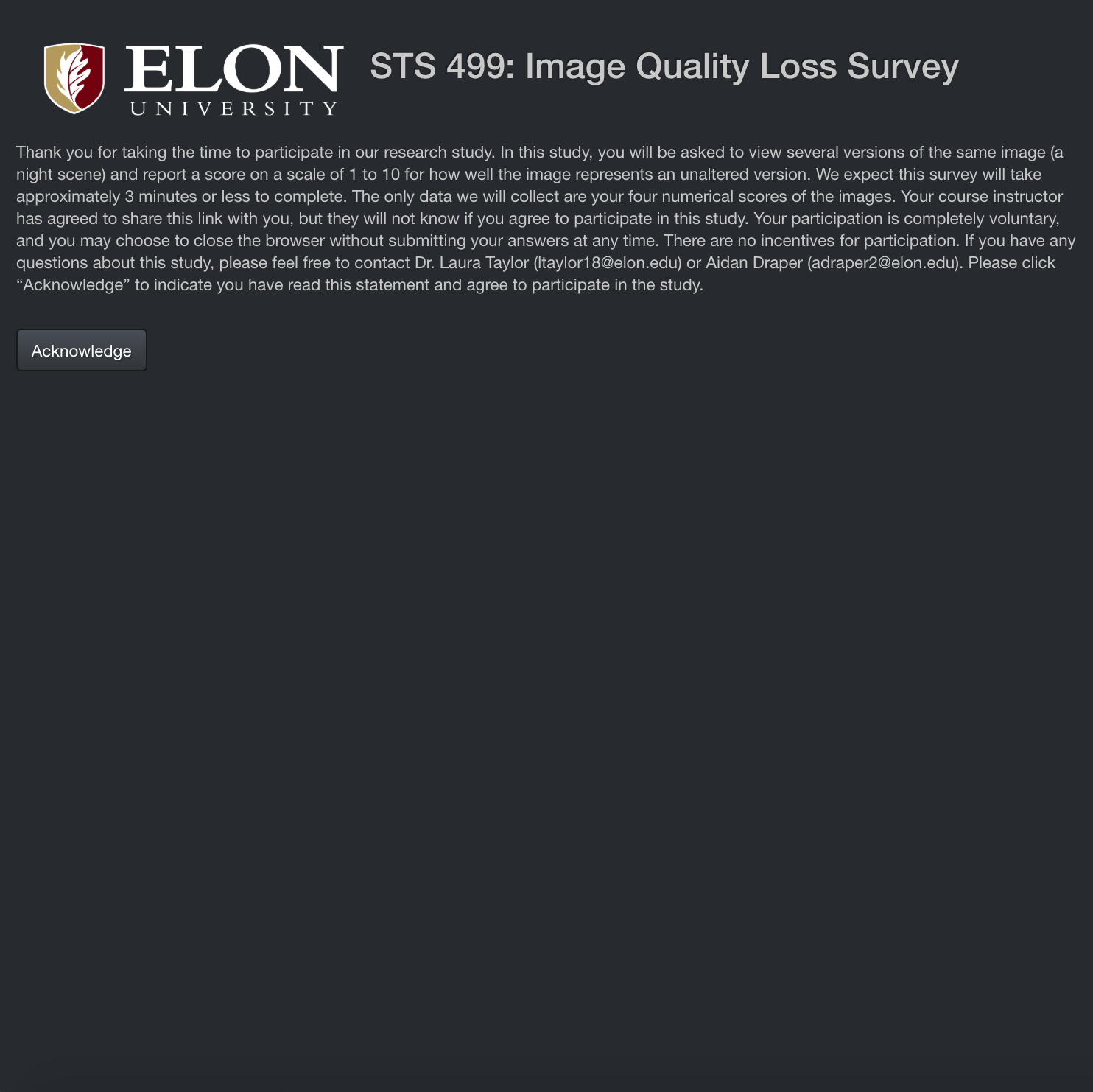}
    \includegraphics[width=0.3\linewidth]{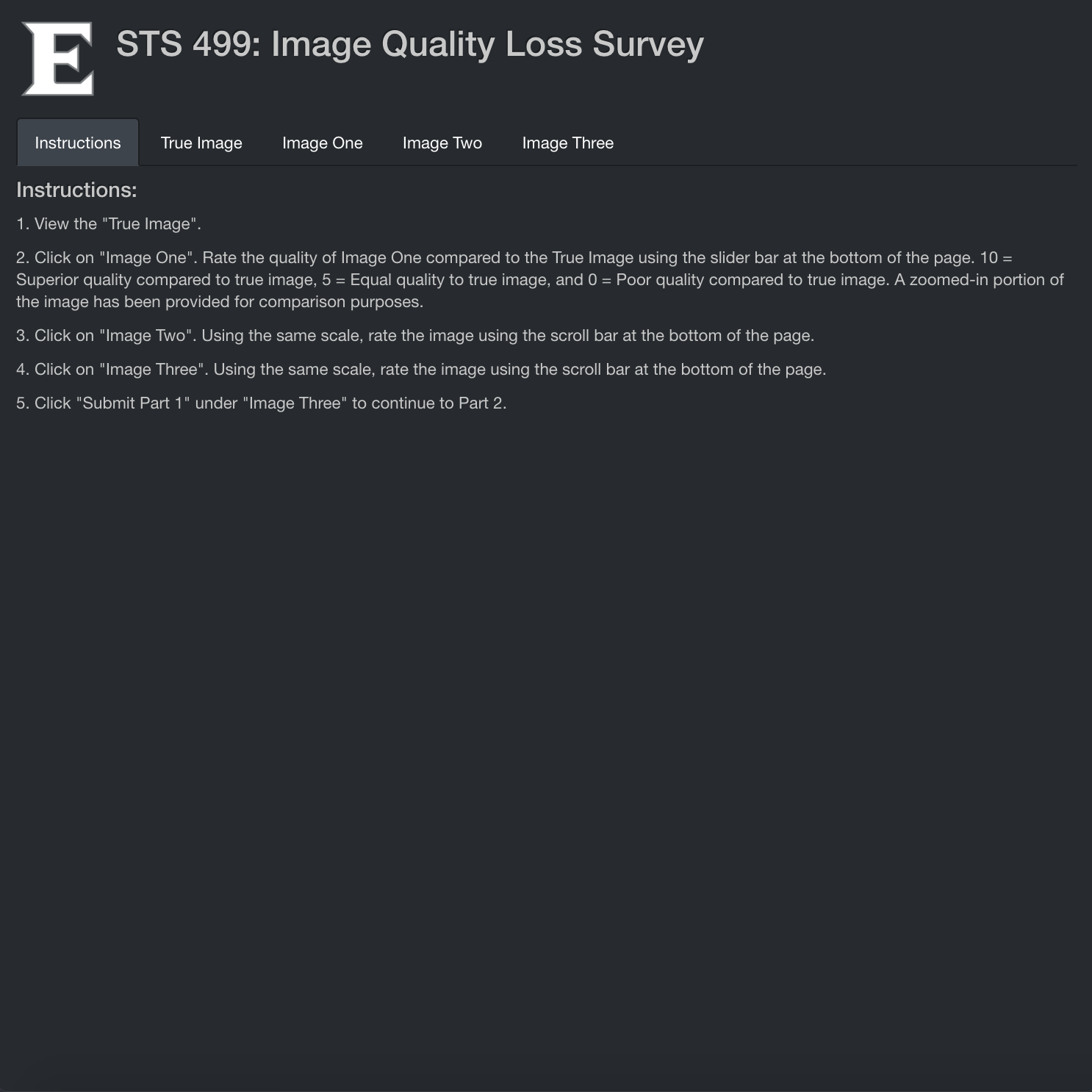}
    \includegraphics[width=0.3\linewidth]{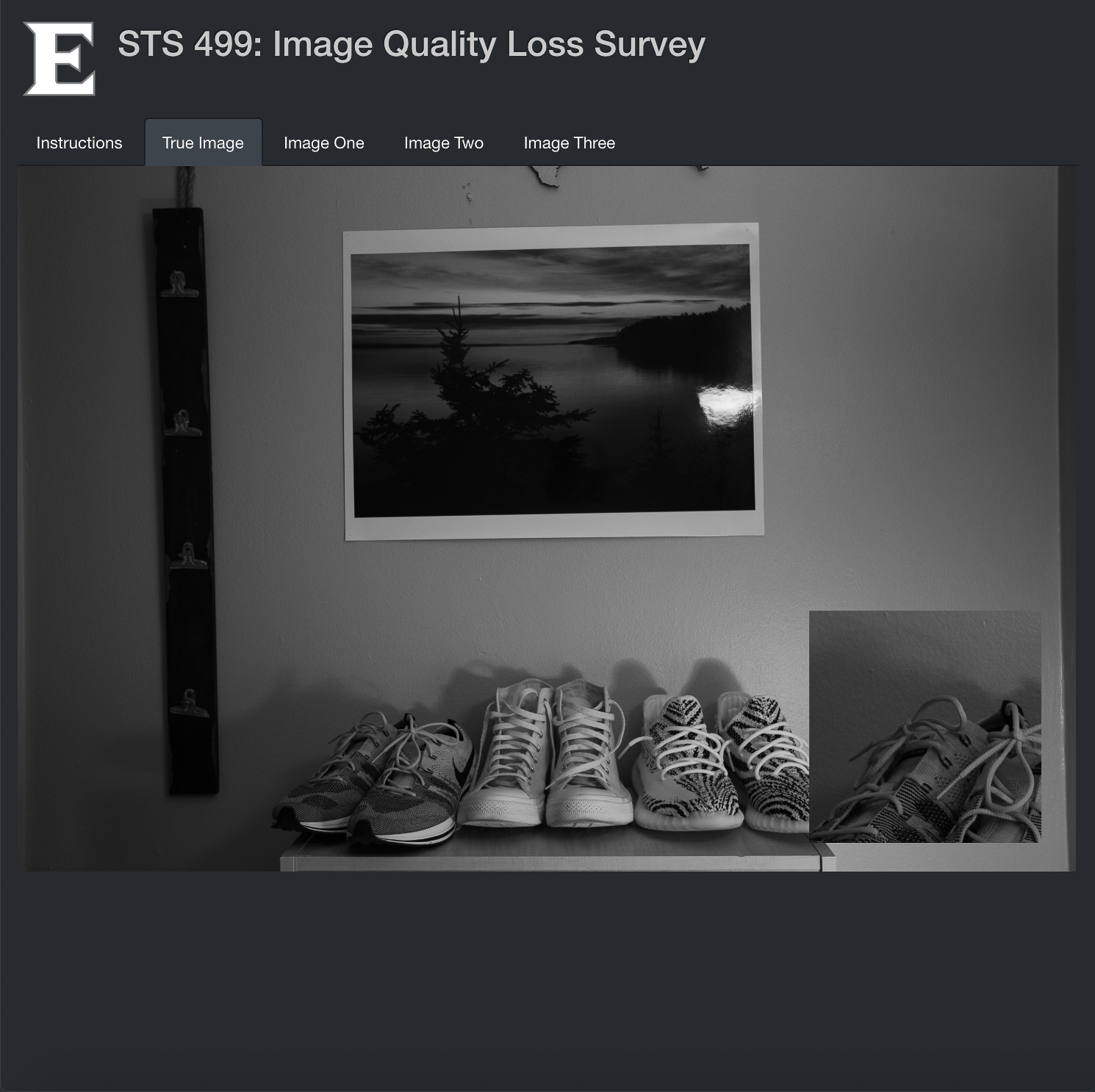}
    \includegraphics[width=0.3\linewidth]{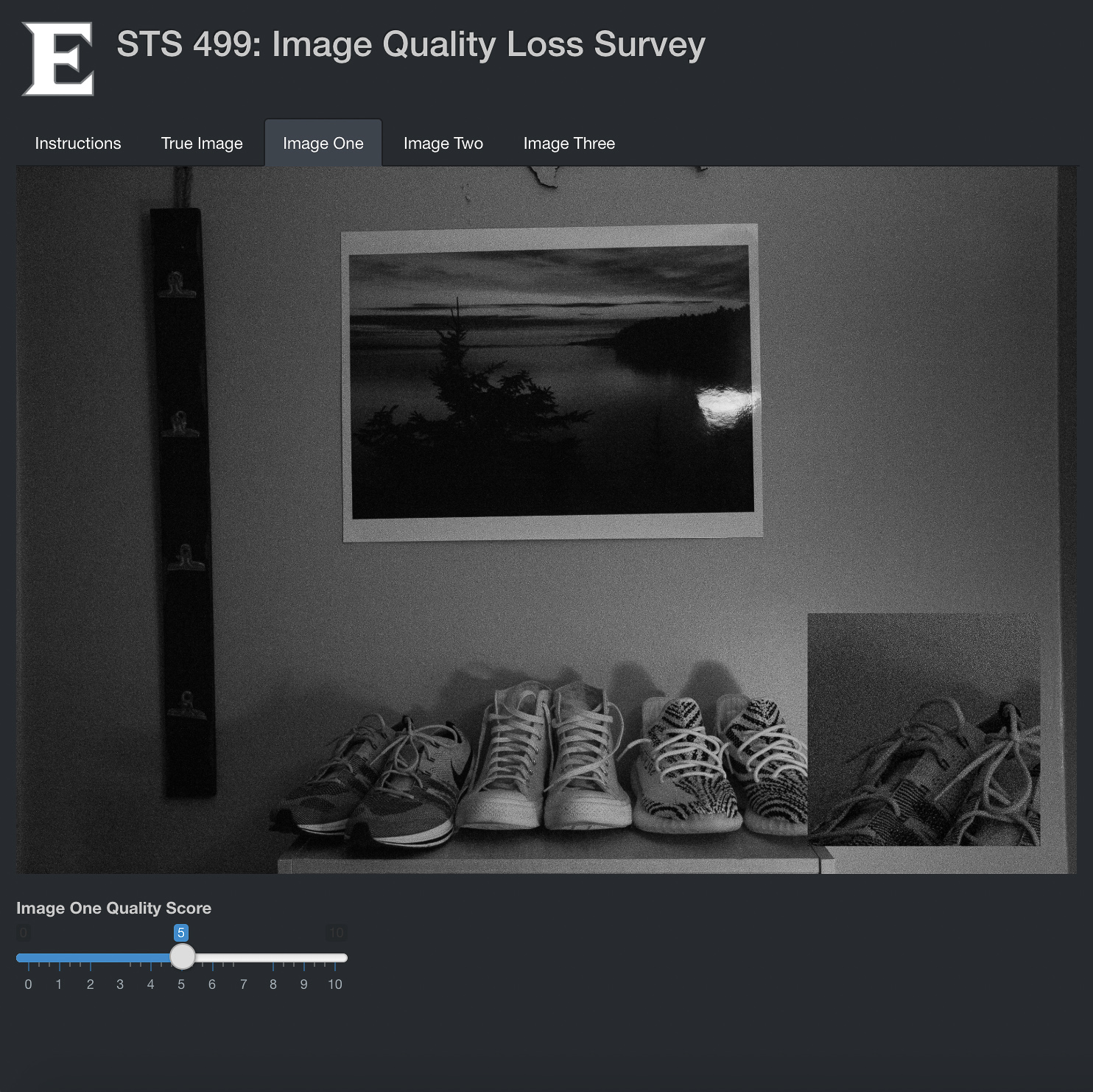}
    \includegraphics[width=0.3\linewidth]{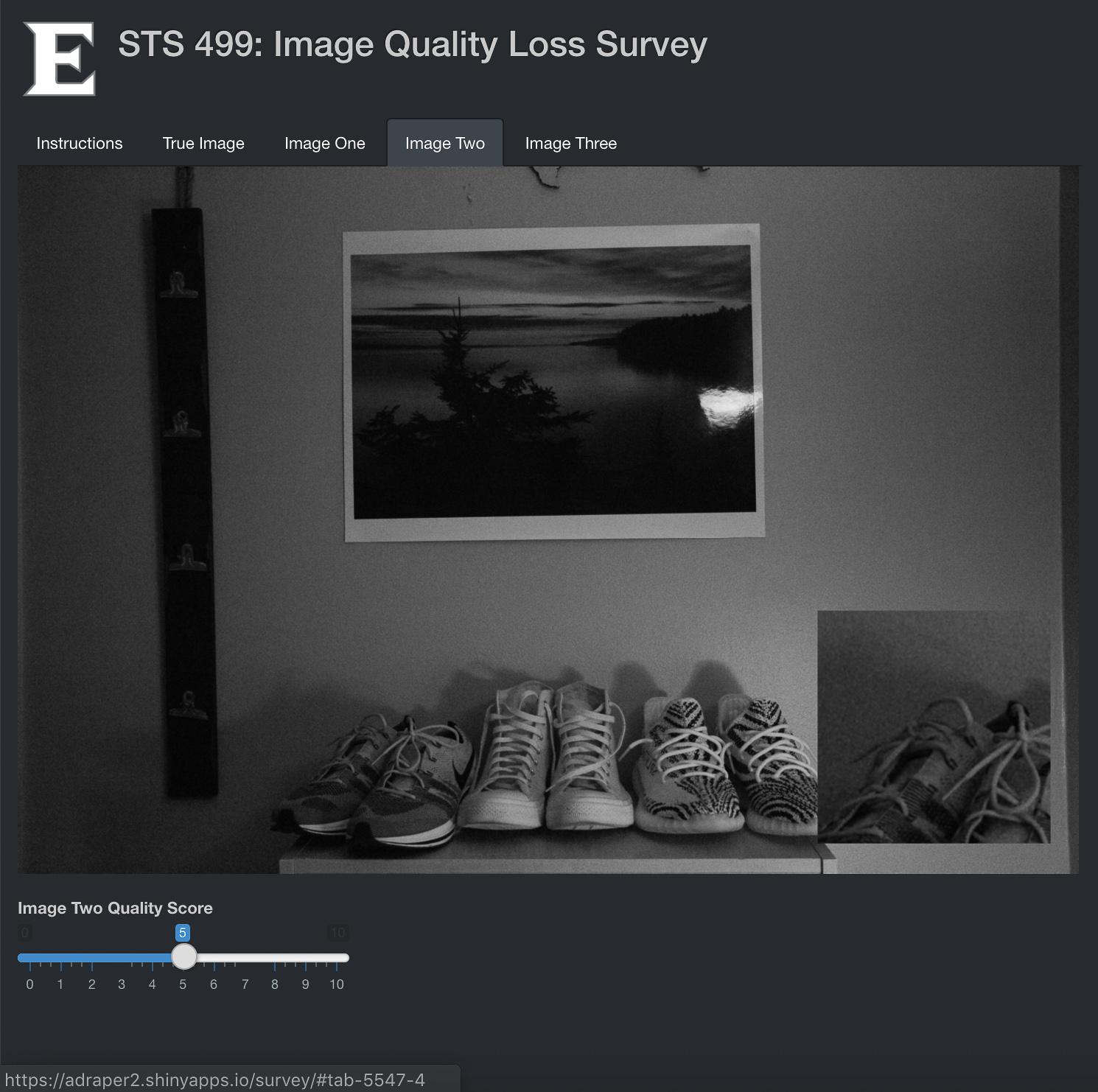}
    \includegraphics[width=0.3\linewidth]{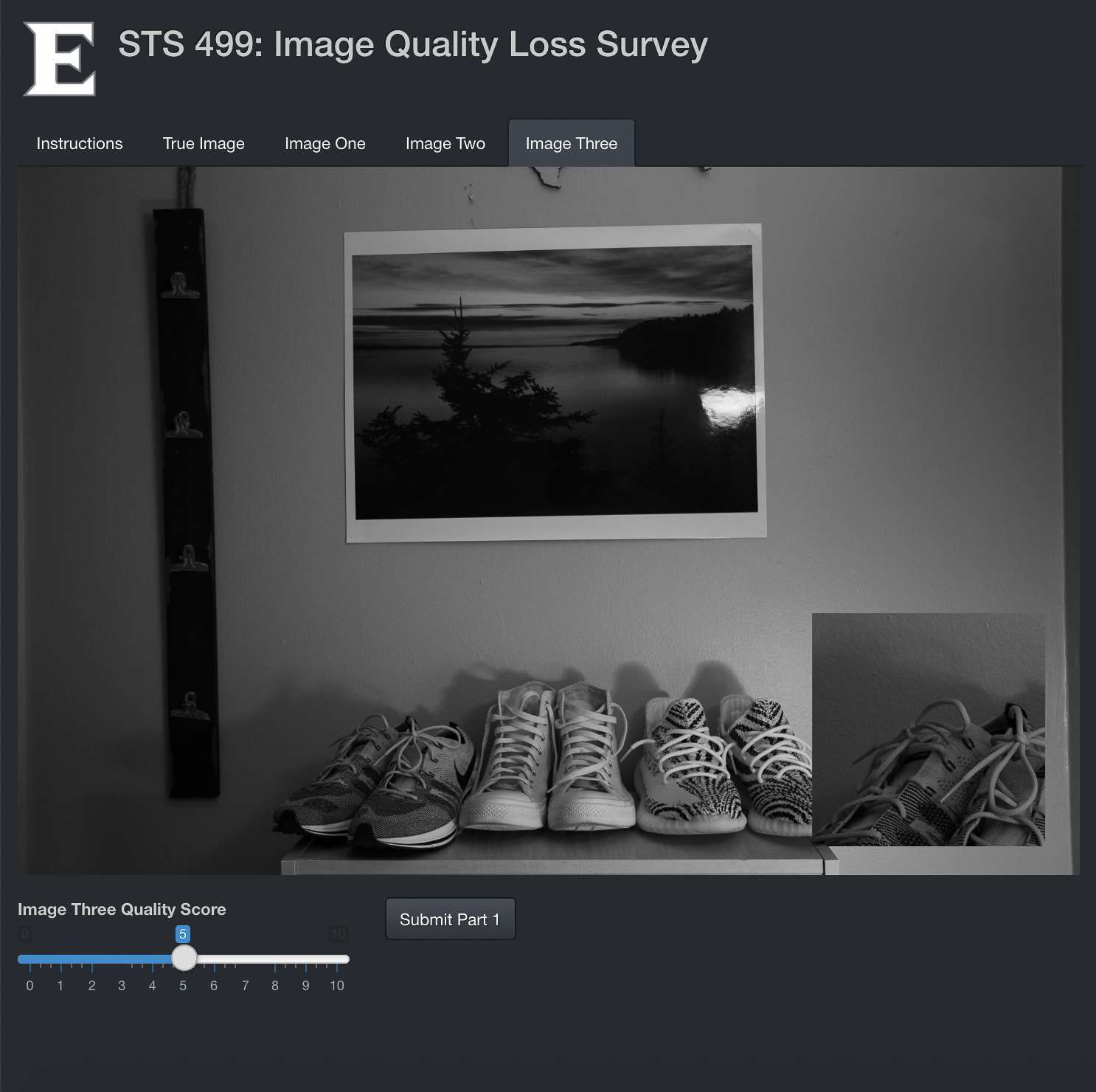}
    \includegraphics[width=0.3\linewidth]{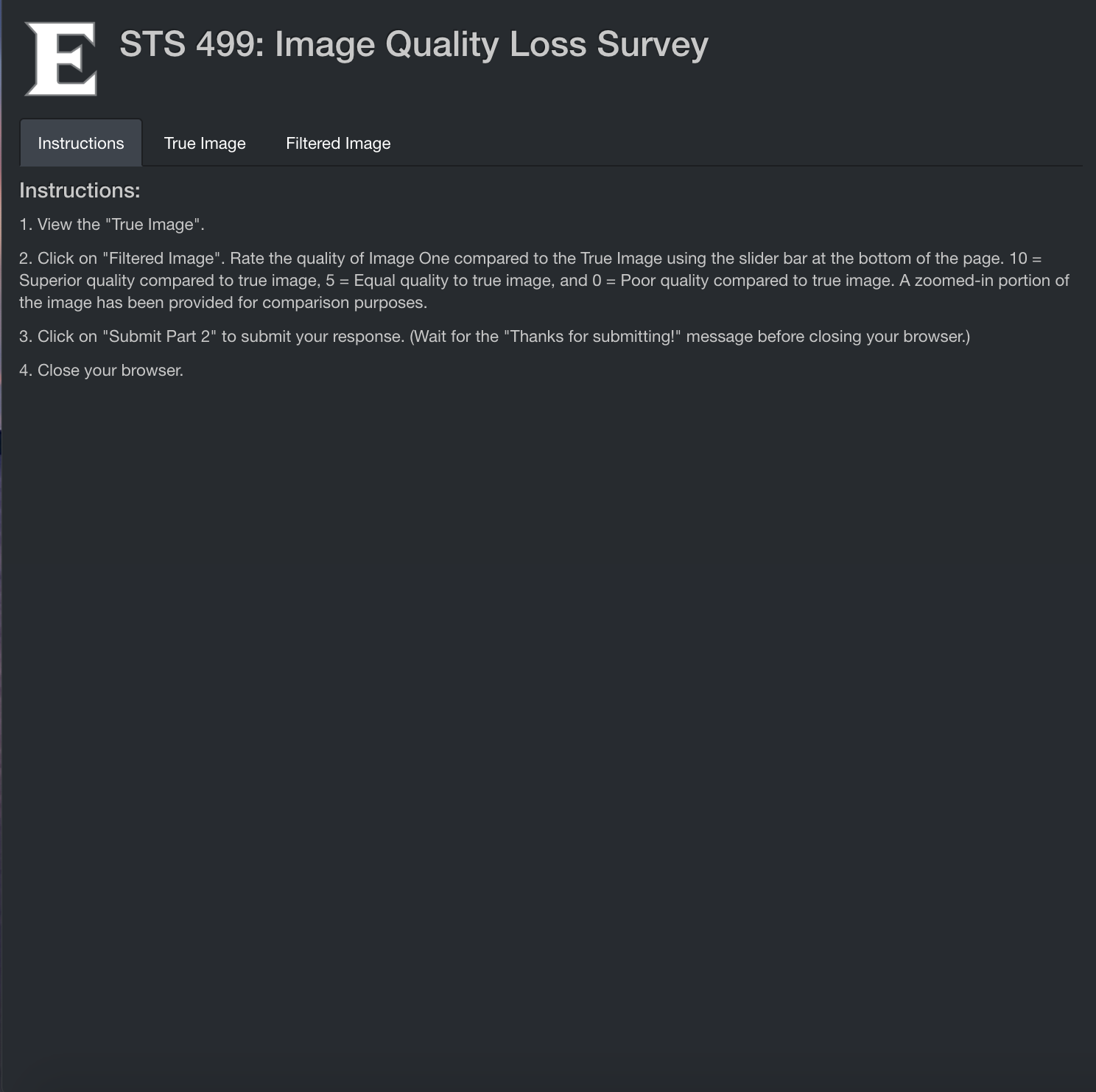}
    \includegraphics[width=0.297\linewidth]{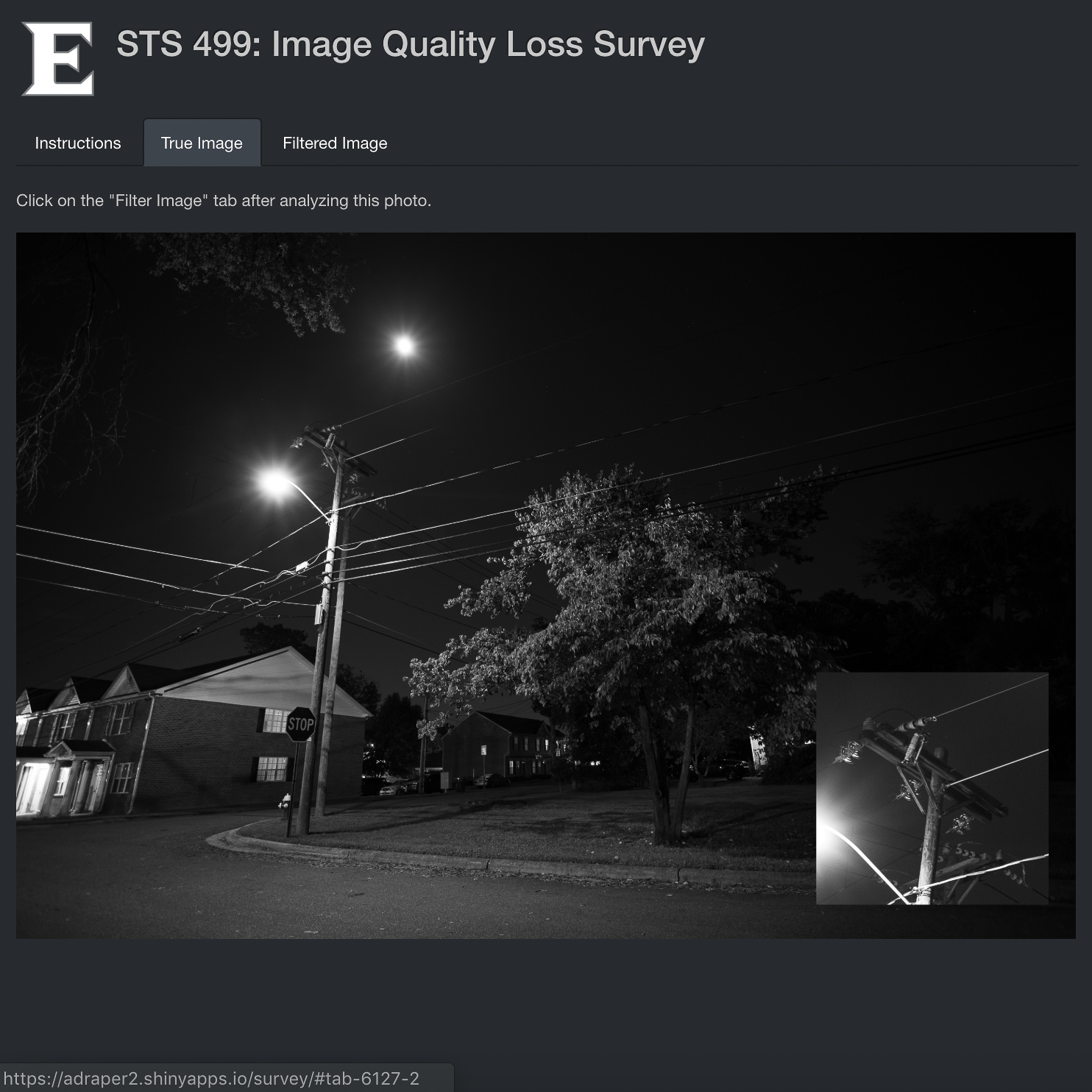}
    \includegraphics[width=0.3\linewidth]{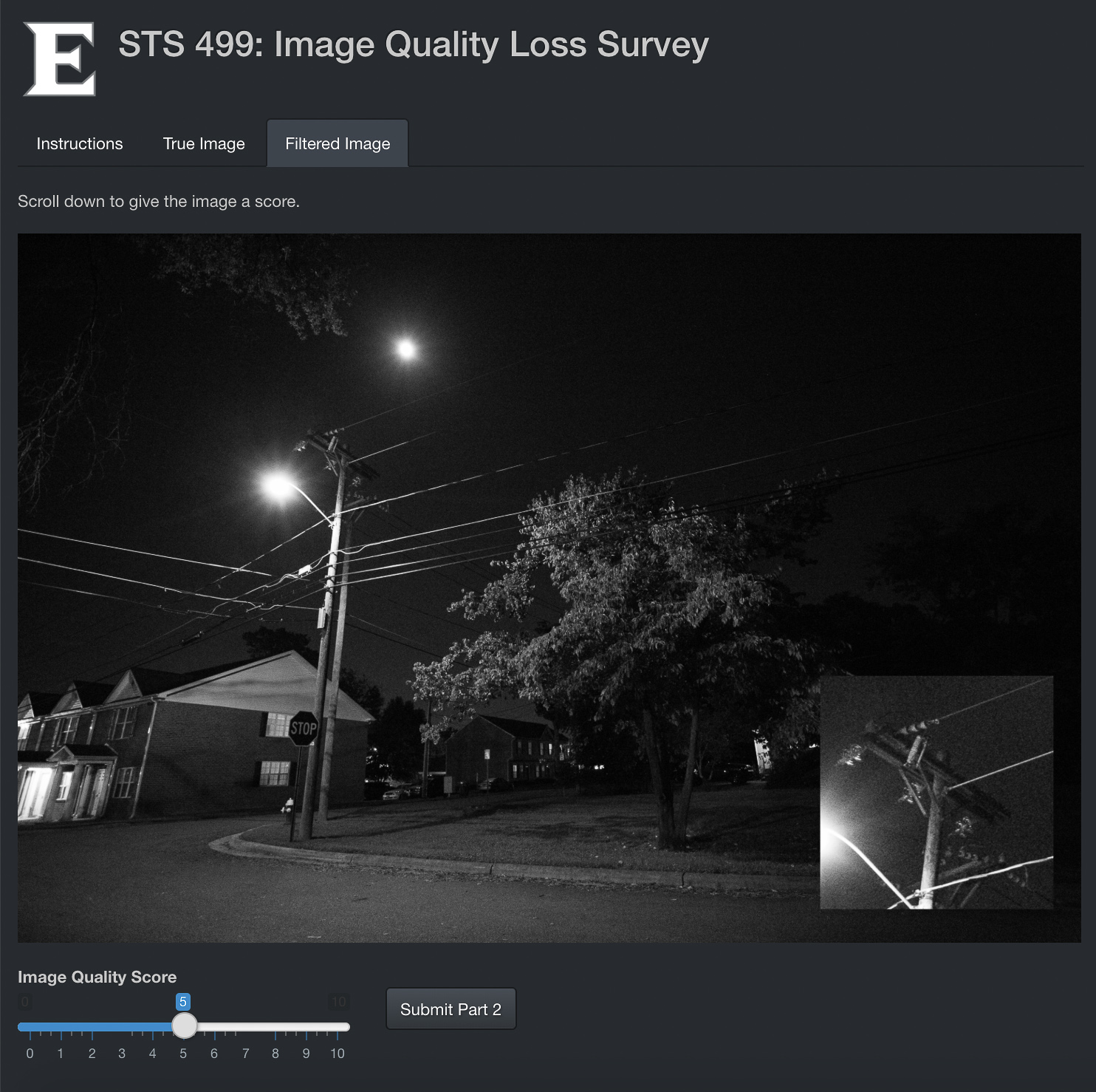}
  \caption{A breakdown of the Shiny application survey.}
  \label{fig:survey}
\end{figure}

\noindent{The experiment process conducted will be expanded on below.}

\subsection{Protocol}
The first step was to compile an original dataset of true and noisy images to test. Although there are image datasets that exist (RENOIR: \citealp{renoir}; Darmstadt: \citealp{darmstadt}), none look to specifically model low-light photography, which was of interest in this study. The camera model used was a full-frame Canon 6D Mark I released back in 2007. It had a USM 17-40mm f/4L Canon lens attached to it. Two photos were taken at each location with greatly varying settings.  First, the original photo would be shot with an ISO of around 200-600 depending on the scene. The aperture was set to the lowest setting in order to let the most light in. The shutter of the camera would be left open for 1 to 5 seconds in order to properly expose the photo at such a low ISO setting. The second photo was shot using the same aperture. However, the ISO setting was increased to just below the camera's maximum light sensitivity settings, which was around 20000 or 25600 ISO, for different scenes to dramatize the amount of noise captured in the image. There are approximately 20 million pixels in each image so noise will not be as apparent when presented regularly.

\begin{figure}[h]
  \centering
    \includegraphics[width=0.4\linewidth]{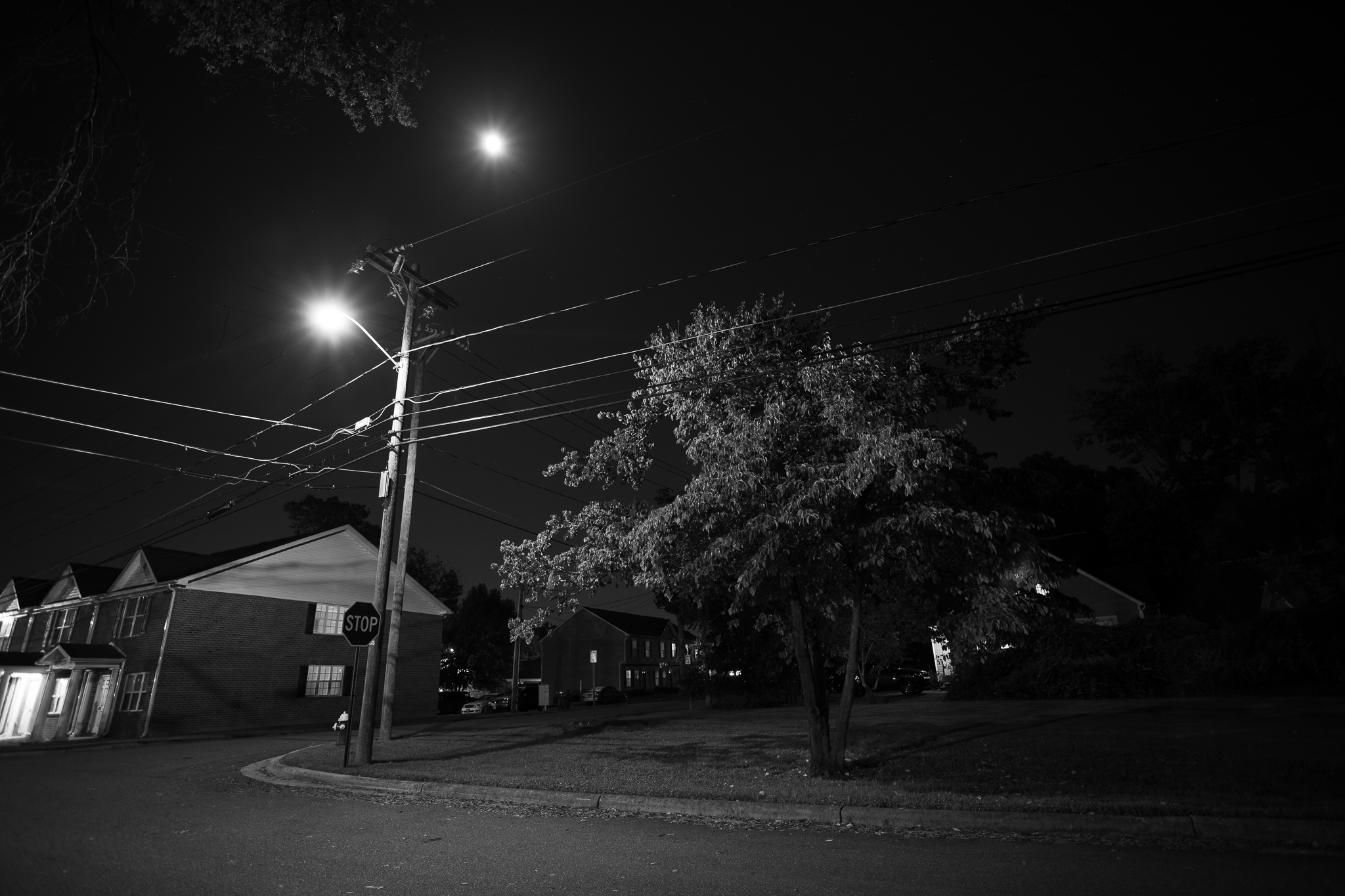}
    \includegraphics[width=0.4\linewidth]{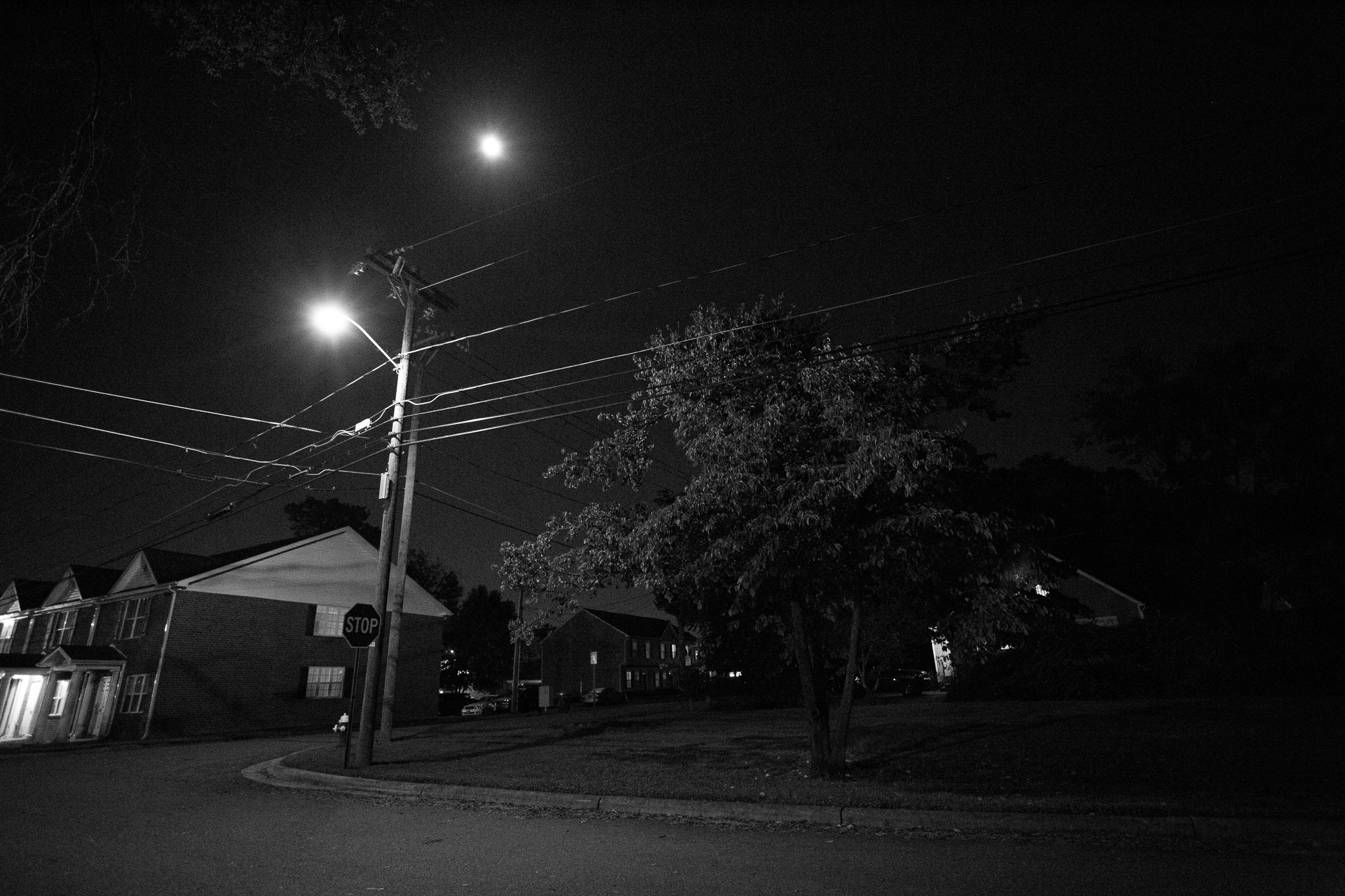}
      \caption{A true state image (left) and a noisy image (right) taken with different camera settings.}
  \label{fig:streetlamp}
\end{figure}

After photos had been captured, they were initially brought into Lightroom where the same grayscale filter was applied to every image. Images were then exported out of their CR2 Canon RAW image formats and into the best possible resolution JPEG format. The Three-by-three Mean,  Non-local Means, and Bilateral Filter python scripts all ran on the noisy image of interest. The unfiltered noisy image was also brought into Adobe Lightroom so that the denoise filter could be applied at the 50\% and 100\% levels. The respective filtered photos were saved and loaded into RStudio. An R script imported the true and filtered images and converted them into matrices. Then, MSE, PSNR, R-Squared, and SSIM were calculated for each filter variation of the same subject matter.

To capture student perceptions of the filtered images, a survey was built using Shiny \citep{shiny} in R. A specific frame was selected so that there was a control in subject matter presented to participants. The subject matter selected for the noisy, filtered and true state images can be seen in \autoref{fig:streetlamp} from earlier. The survey began by presenting respondents with a noiseless photo and then, asking them to rate three additional photos of the same content, but with varying degrees of image quality. The first training photo presented was the unfiltered noisy photo. The second training photo was a Non-local Means filtered photo. The third training photo was the same photo as the noiseless photo. These three scores were meant to serve as an indication of an inability to identify noise, or image quality, in respondents while also taking into account that respondents will have different metrics for image quality, naturally. Once the initial training portion was complete, respondents were asked to rate a final photo about its image quality on a scale of 1-10 in comparison to its respective noiseless image. A score of one represented an image that had far inferior quality as compared to the original. A score of five indicated equal image quality to the respective noiseless image. Lastly, a score of ten meant that the respondent felt that the photo had far superior image quality to the noiseless photo.

Once the respondent had selected scores for those three training images, they proceeded to the images of interest. Again, the noiseless image was presented first and then, they were asked to compare a filtered photo that had been randomly assigned to them to the noiseless photo of the same subject matter. 

\begin{figure}
  \includegraphics[width=0.7\linewidth]{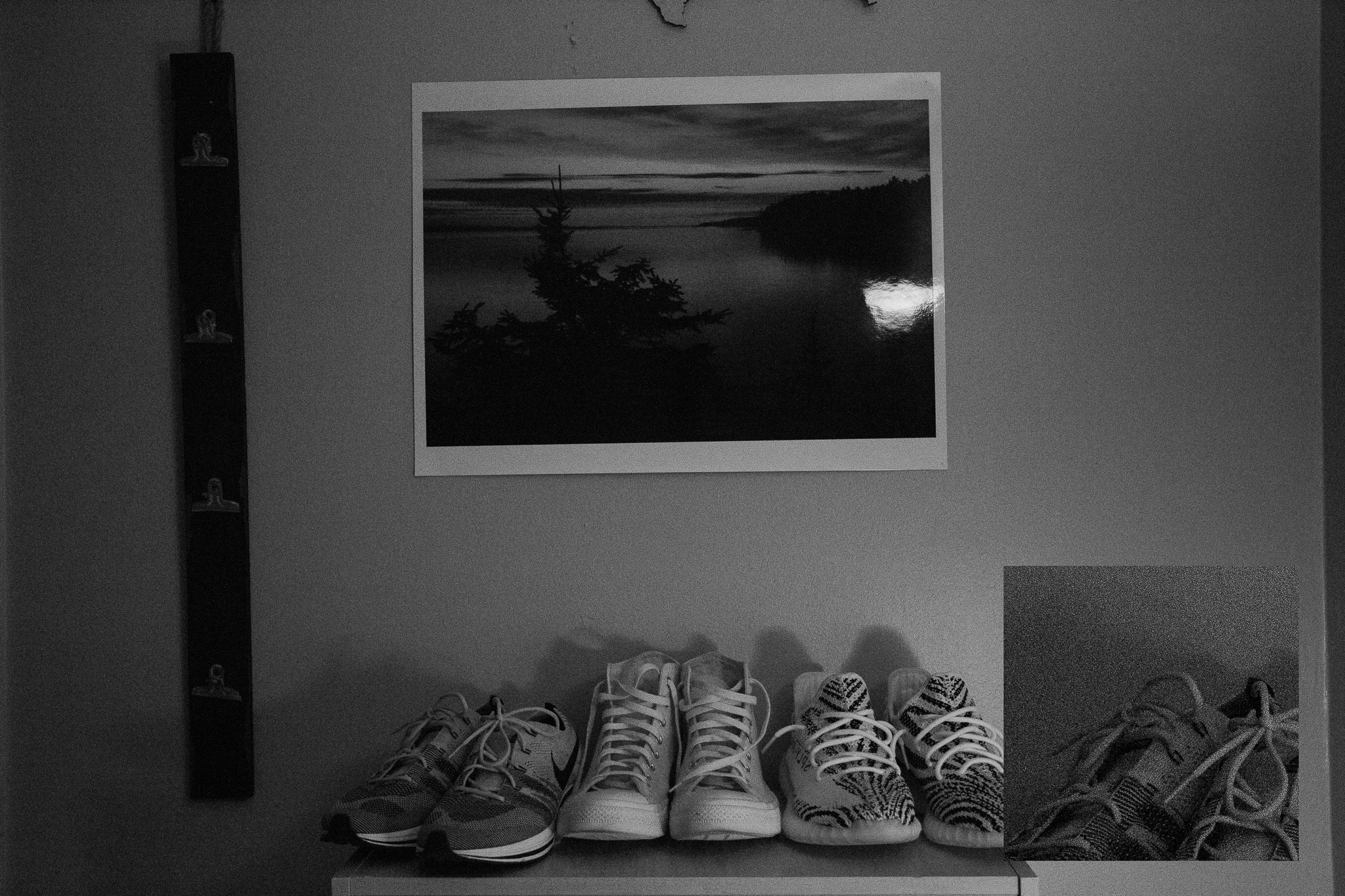}
  \caption{An example of an image with a blown-up region (in the lower right corner).}
  \label{fig:blownup}
\end{figure}

To emphasize differences in images, a blown-up region of each image was included in the bottom right-hand corners. These blown-up regions were 600-by-600 pixel grids from sections of the photo that included an edge. In Photoshop, these regions were enlarged to approximately 1200-by-1200 pixels. An example of this layout can be seen in \autoref{fig:blownup}. After the participant rated the last image, scores were sent to a Google Sheet.

\section{Experiment Results}
\begin{figure}[h]
  \centering
    \includegraphics[width=0.3\linewidth]{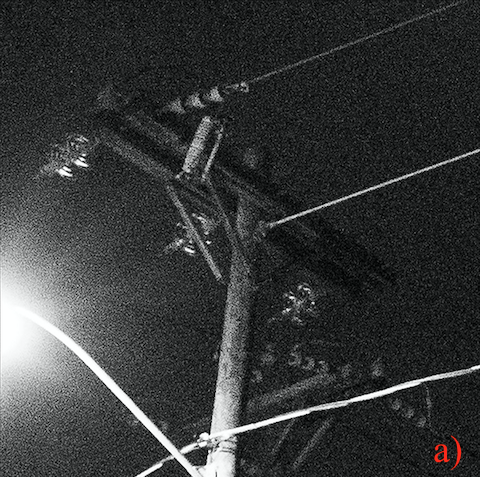}
    \includegraphics[width=0.3\linewidth]{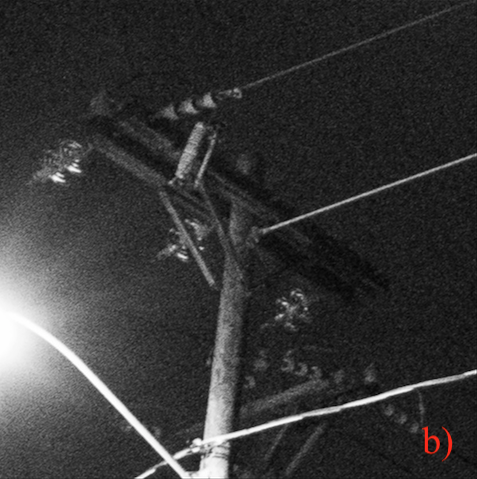}
    \includegraphics[width=0.3\linewidth]{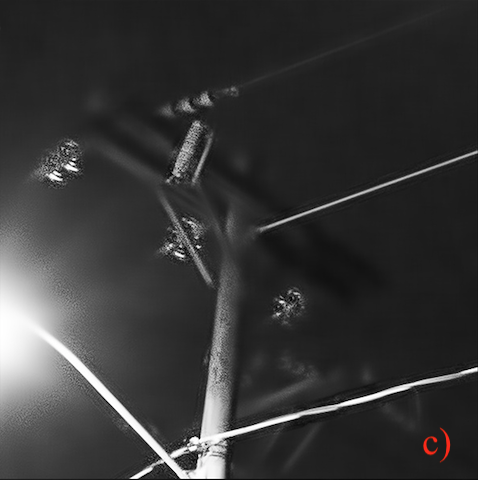}
    \includegraphics[width=0.3\linewidth]{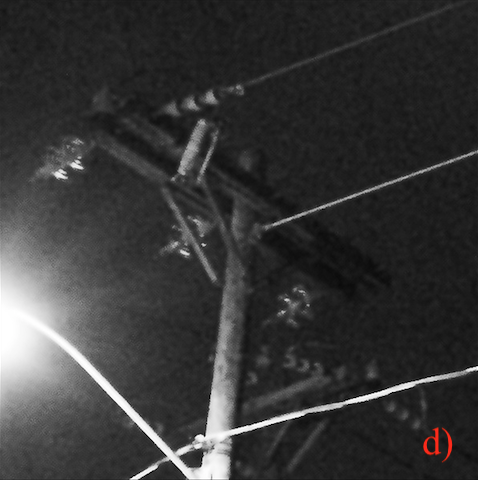}
    \includegraphics[width=0.3\linewidth]{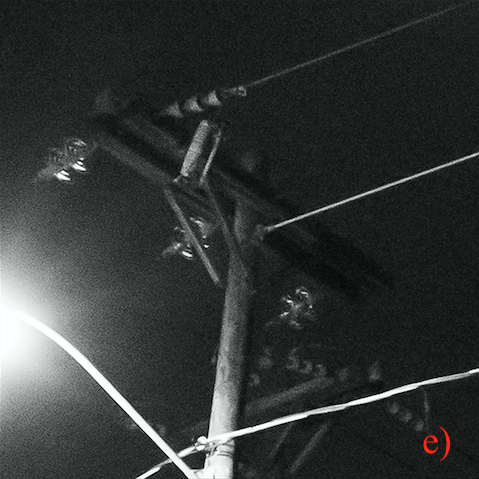}
    \includegraphics[width=0.3\linewidth]{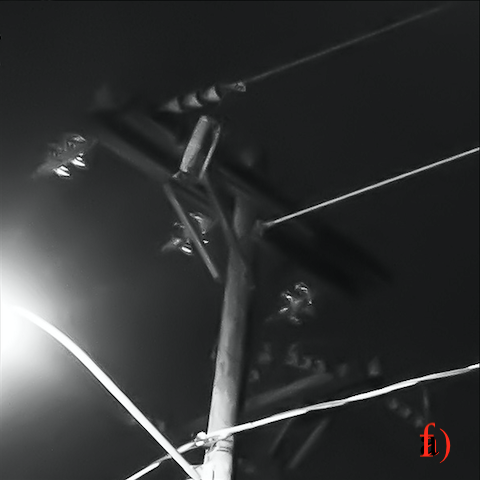}
  \caption{A grid of 600 by 600 pixel regions of an unfiltered image and its filtered images. The regions are ordered as follows: Noisy (a), Three-by-three Mean (b), Non-local Means (c), Bilateral (d), Adobe 50\% (e), and Adobe 100\% (f).}
  \label{fig:images}
\end{figure}

Visual interpretations of image quality were made for each processed photo prior to running benchmark tests. Zoomed-in regions of each filtered photo were produced so that differences between methods may be easier to compare in this report. The resultant images are shown next to the original noisy image in \autoref{fig:images}. Note that this paragraph is subjective based on the visual opinions of the authors. The authors believe that, visually, the Adobe 50\% (e) and Bilateral Filter (d) produced the most compelling filtered images. This decision can be justified by their preservation of fine-detail, as well as decent edge preservation, when compared to the original noisy image (a). These images appear less washed due to a lack of blurring. The Three-by-three Mean Filter (b) also produced less blurring, but noise was far more present in this image, which is most visible in the gradient of light originating from the street lamp. Noise is less present in the light gradient for images $d$ and $e$ of \autoref{fig:images}. Although images $c$ and $f$ also show less noise in the streetlamp light gradient, the smoothness of the gradient is an indicator of blurring, which is more obvious when contrasting the top of the telephone pole across filter methods. Images $c$ and $f$ have sacrificed edge preservation,  arguably just as an important of a factor in evaluating the clarity of photos, for noise reduction. The least compelling result was from the Non-local Means Filter (c), which did not preserve edges or fine details well. The filtered image has no noise, but has also destroyed image clarity in the process, as the top of the telephone pole appears to blend into the background. The Three-by-three Mean Filter (b) appeared to be least effective at removing noise, but detail and edges have been preserved due to the lack of blurring. The Adobe 100\% (f) produced the most blurring, yet noise is nonexistent in its resultant image.

% keep working on this paragraph (WHERE TO START)
Benchmark scores were calculated for every filter method, as well as the noisy photo, in comparison to a noiseless image. With the exception of the SSIM scores for the Adobe 50\% and Adobe 100\% filters, all methods scored higher than the unfiltered noisy image in the benchmark tests, which means almost every filter method showed some degree of improvement in image reconstitution. The debatable filters are Adobe 50\% and Adobe 100\%, who scored well in every test except for SSIM. PSNR, MSE and $R^2$ scored images similarly and ranked methods from most improved to least improved as Bilateral, Adobe 100\%, Non-local Means, Three-by-three Means, and Adobe 50\%. SSIM offered a different conclusion that declared Adobe 50\% and Adobe 100\% as less structurally similar to the noisy photo, which would mean that the benchmark believes image reconstitution has worsened after filtering. The Bilateral Filter still outperformed every other filter (SSIM=0.8577, PSNR=47.8103, MSE=264.8607, $R^2$=0.7217, Run time=0.1186 seconds), which declares it as the quickest and most effective filtering method in this study. \autoref{table:benchmark} reports the five benchmark scores for these images for reference. 

\begin{table}[ht]
\caption{Benchmark Results for the Street Lamp Image.} % title of Table
\centering % used for centering table
\begin{tabular}{|c|r|r|r|r|r|} % centered columns (4 columns)
\hline %inserts double horizontal lines
 & \multicolumn{1}{c|}{\textbf{SSIM}} & \multicolumn{1}{c|}{\textbf{PSNR}} & \multicolumn{1}{c|}{\textbf{MSE}} & \multicolumn{1}{c|}{$\mathbf{R^2}$} & \multicolumn{1}{c|}{\textbf{Run time (s)}} \\ [0.5ex] % inserts table
%heading
\hline % inserts single horizontal line
\textbf{\textit{Unfiltered}} & 0.7520 & 41.7691 & 530.4268 & 0.4426 & N/A  \\
\textbf{3x3 Mean} & 0.8432 & 46.8833 & 294.3851 & 0.6732 & 91.9112 \\ 
\textbf{Non-local} & 0.8445 & 46.9004 & 293.8056 & 0.6913 & 20.5480 \\
\textbf{Bilateral} & 0.8577 & 47.8103 & 264.8607 & 0.7217 & 0.1886 \\
\textbf{Adobe 50\%} & 0.7111 & 46.3890 & 311.6243 & 0.6732 & 0.5532 \\
\textbf{Adobe 100\%} & 0.7245 & 47.5747 & 271.8604 & 0.7143 & 1.3872 \\ [1ex] % [1ex] adds vertical space
\hline %inserts single line
\end{tabular}
\label{table:benchmark} % is used to refer this table in the text
\end{table}

% TALK ABOUT BENCHMARK RESULTS MORE HERE

In total, there were 89 responses received from the survey as far more students chose not to respond than to respond to the survey. However, it is important to note that three data points had to be removed from the experiment. For at least one of the training images, the respondent marked a zero value for a training score, which is nonexistent on the range of possible values. This is likely from a malfunction in collecting data in Shiny Apps or sending the data using the `googlesheets' R library \citep{googlesheets}. The descriptive statistics of image quality scores, with the bad data points removed, are visible in \autoref{table:desc_survey}.

\begin{table}[ht]
\caption{Summary of survey responses based on filter presented to respondent. Aside from sample size, units of measurement are points on a scale from 1 (inferior quality) to 10 (superior quality).} % title of Table
\centering % used for centering table
\begin{tabular}{|c|r|r|r|r|} % centered columns (4 columns)
\hline
 & \multicolumn{1}{c|}{\textbf{n}} & \multicolumn{1}{c|}{$\mathbf{\bar{x}}$} & \multicolumn{1}{c|}{$\mathbf{\Tilde{x}}$} & \multicolumn{1}{c|}{\textbf{\textit{s}}}\\ [0.5ex] 
 % inserts table
%heading
\hline % inserts single horizontal line
\textbf{3x3 Mean} & 15 & 3.6 & 4.0 & 0.99  \\ 
\textbf{Non-local} & 18 & 4.1 & 4.0 & 1.76 \\
\textbf{Bilateral} & 18 & 4.0 & 3.5 & 1.91 \\
\textbf{Adobe 50\%} & 18 & 4.4 & 4.0 & 2.04 \\
\textbf{Adobe 100\%} & 17 & 5 & 4.0 & 2.42 \\ [1ex] % [1ex] adds vertical space
\hline %inserts single line
\end{tabular}
\label{table:desc_survey} % is used to refer this table in the text
\end{table}

\noindent{The Adobe 100\% Filter had the largest mean image quality score ($\bar{x}_{\text{Adobe 100\%}} = 5.0$), followed by the Adobe 50\% Filter ($\bar{x}_{\text{Adobe 50\%}} = 4.4$). They also had the largest spread to their image scores ($s_{\text{Adobe 100\%}}=2.42$, $s_{\text{Adobe 50\%}}=2.04$). Image scores for the Adobe 100\% were skewed right, which apparent by a sample mean larger than the sample median. This characteristic tells us that a few respondents scored this filter group far better than others did. The filter group that received the most criticism, but most precision, was the Three-by-three Mean Filter ($\bar{x}_{\text{3x3 Mean}} = 3.6$, $s_{\text{3x3 Mean}}=0.99$). Surprisingly, the Bilateral Filter received lower scores for image quality ($\bar{x}_{\text{Bilateral}}$=4.0, $\Tilde{x}_{\text{Bilateral}}$=3.5), even though it outperformed every filter during benchmark testing ($\text{SSIM}_{\text{Bilateral}}$=0.8577, $\text{PSNR}_{\text{Bilateral}}$=47.8103). Note that the lack of responses, as well as bad data points, lead to uneven sample sizes between filter groups.}

\begin{figure}[ht]
  \centering
    \includegraphics[width=0.8\linewidth]{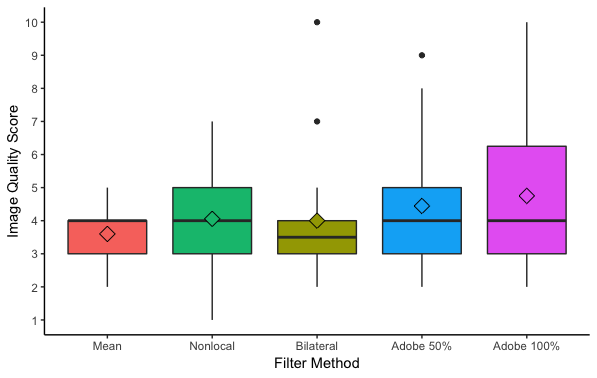}
  \caption{Boxplots of the participants' image quality perception score distributions by filtering method.}
  \label{fig:box}
\end{figure}

The distribution of image quality scores by filter groups is visualized in \autoref{fig:box}. Filter group image score sample means remain fairly similar. Shockingly, all filter method image score distributions, with the exception of the Three-by-three Mean filter, are being skewed by a few scores that would imply a superior image quality to the filtered photo's noiseless counterpart. The programmers of these algorithms would surely be satisfied knowing these select respondents' sentiments, although, there is likely another conclusion that could be drawn from this observation. Adobe 100\% received almost every image quality score, giving it the largest range, suggesting that respondents' opinions differed the most in judging this image. The overlap in interquartile ranges suggests a non-significant p-value from the ANOVA test.

\begin{table}[ht]
\caption{An ANOVA table for the image quality perception scores.} % title of Table
\centering % used for centering table
\begin{tabular}{|c| rrrrr|} % centered columns (4 columns)
\hline %inserts double horizontal lines
& \multicolumn{1}{c}{\textbf{Df}} & \multicolumn{1}{c}{\textbf{SSE}} & \multicolumn{1}{c}{\textbf{MSE}} & \multicolumn{1}{c}{\textbf{F-score}} & \multicolumn{1}{c|}{\textbf{P-value}} \\ [0.5ex]
%heading
\hline % inserts single horizontal line
\textbf{Treatment} & 4 & 18.36 & 4.590 & 1.269 & 0.289 \\ % inserting body of the table
\textbf{Error} & 81 & 292.99  & 3.617 & &  \\ [1ex]
\hline
\textbf{Total} & 85 & 311.35 & & & \\ % [1ex] adds vertical space
\hline %inserts single line
\end{tabular}
\label{table:anova_table} % is used to refer this table in the text
\end{table}

As apparent in \autoref{table:anova_table}, the one-way analysis of variance test yielded insufficient evidence of a difference in the population mean image quality score across the noise filtering methods tested (F(4,81) = 1.269, p = 0.289). More of the residual error is found within individual methods. This is likely due to the great variation in scores received from respondents for individual methods, which is shown in \autoref{fig:box}. With the exception of the Three-by-three Mean Filter, the remaining groups had standard deviations of approximately two, according to \autoref{table:anova_table}, implying that the scores were fairly spread around the mean and that respondents often did not share the same opinions about the image quality of each filtered photo. In testing assumptions, it became apparent that there was one outlier with a Studentized residual greater than three ($s_{e_{21}} = 3.46$). There still was insufficient evidence to reject the null hypothesis under an alpha level of 0.05 when testing the data with no outliers. Log and square root transformations were tested. The residual error had a better spread across fitted values, but again, the conclusion about the null hypothesis would not change.

\begin{figure}[h]
  \centering
    \includegraphics[width=0.9\linewidth]{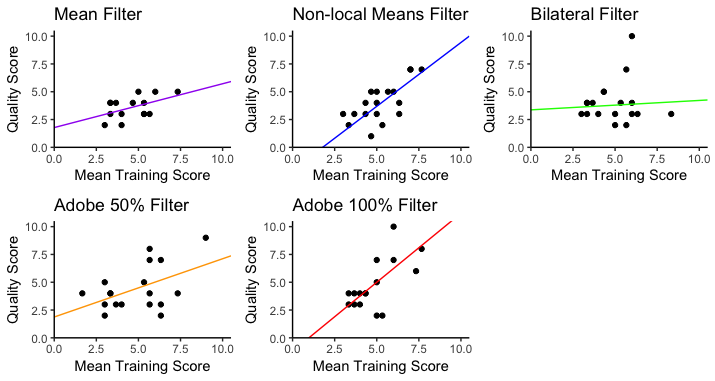}
  \caption{The unequal slopes ANCOVA lines fitted to scatterplots by image filter.}
  \label{fig:un_ancova}
\end{figure}

When analyzing the image quality scores, it became apparent that the initial training scores received could be indicators of an unexplained variance on the final image score received due to a similar trends in variation in training score distributions to the final image score distributions. As a result, an ANCOVA test was conducted to determine if removing the potential effect of training scores, or individual respondent image quality subjectivity, would lead to sufficient evidence to reject the null hypothesis that each filter group had the same population mean image quality score. Additionally, an analysis of covariance could explain how differences in training scores affected the final image scores received between filter methods. It was suggested that the average training score of the initial three images could be an influence on, and explain uncontrolled variation in, respondents' image quality scores. The unequal slopes model for the relationship between mean training score and image quality score is graphed in \autoref{fig:un_ancova}.

% talk about ANCOVA and provide formula with indicator variables for both equal and unequal variance

There was slightly insufficient evidence of an interaction effect between the mean training score and the final image quality score in the unequal slopes model (F(4,81) = 2.40, p = 0.057). The least parallel slope to the rest of the filter methods was the slope for the Bilateral Filter. This result technically validates the use of the equal slopes model, but 95\% confidence intervals were calculated for both models because of how small the p-value was. The 95\% confidence intervals are reported in \autoref{table:un_ancova_ci}.

\begin{table}[ht]
\caption{The unequal slopes 95\% Confidence Interval estimates of the mean training score on image score slopes for each filter method.} % title of Table
\centering % used for centering table
\begin{tabular}{|c| r|r|r|} % centered columns (4 columns)
\hline %inserts double horizontal lines
& \multicolumn{1}{c|}{\textbf{Slope Estimate}} & \multicolumn{2}{c|}{\textbf{95\% C.I.}}  \\ [0.5ex]
%heading
\hline
\textbf{3x3 Mean} & 0.3943 & -0.6183 & 1.4069  \\ 
\textbf{Non-local} & 1.1482 & 0.2251 & 2.0712 \\
\textbf{Bilateral} & 0.0847 & -0.8195 & 0.9888 \\
\textbf{Adobe 50\%} & 0.5238 & -0.2800 & 1.3277 \\
\textbf{Adobe 100\%} & 1.2403 & 0.5791 & 1.9015 \\ [1ex]
\hline %inserts single line
\end{tabular}
\label{table:un_ancova_ci} % is used to refer this table in the text
\end{table}

\noindent{According to the unequal slopes model, two filter methods had significantly positive 95\% confidence intervals for the estimated slope of mean training image score on image quality score (Adobe 100\%: 95\% CI = [0.5791, 1.9015], Non-local: [0.2251, 2.0712]). The unequal slopes model equation that estimates the score received on the final image based on filter method and mean score of training images (t) is given by:}

\begin{equation}
  \begin{array}{l}
    \Hat{y}=-1.20 + 2.98(I_{\text{3x3 Mean}}) + 4.57(I_{\text{Bilateral}}) - 0.85(I_{\text{Non-local}}) + 3.08(I_{\text{Adobe 50\%}}) + \\ 
    1.24*t - 0.85(I_{\text{3x3 Mean}}*\text{t}) - 1.16(I_{\text{Bilateral}}*\text{t}) - 0.09(I_{\text{Non-local}}*\text{t}) - 0.72(I_{\text{Adobe 50\%}}*\text{t})
  \end{array}
\end{equation}

\noindent{where $I_{i}$ represents an indicator random variable for a filter and $t$ indicates the slope of model that fits the mean training score and the predicted image quality score, $\Hat{y}$. The model is adjusted around the linear equation for the Adobe 100\% Filter, which is why there is no indicator random variable for that filter.}

\begin{figure}[h]
  \centering
    \includegraphics[width=0.9\linewidth]{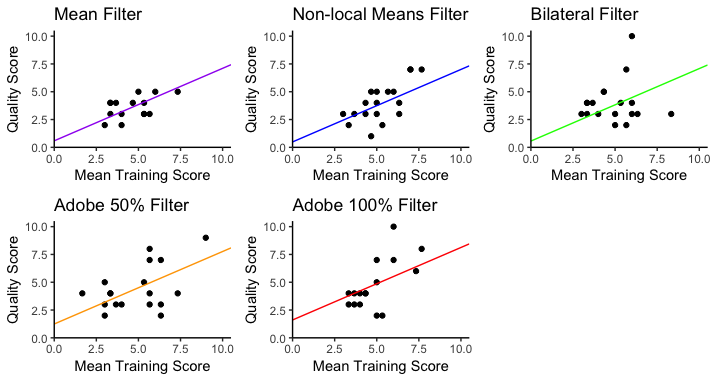}
  \caption{The equal slopes ANCOVA lines fitted to scatterplots by image filter.}
  \label{fig:eq_ancova}
\end{figure}

Since there was marginally insignificant evidence of unequal slopes at the alpha level of 0.05 (p = 0.057), the equal slopes model was also fitted. This model is depicted in \autoref{fig:eq_ancova} and provides a common estimate of the slope associated with mean training score (Estimate = 0.6519, 95\% CI = [0.3739, 0.9299]), which is significantly positive. The equal slopes model equation to estimate the training image effect on the predicted final image scores across all filter methods is given by:

\begin{equation}
    \Hat{y}= 0.65(t) + 1.61 - 1.02(I_{\text{3x3 Mean}}) - 1.04(I_{\text{Bilateral}}) - 1.12(I_{\text{Non-local}}) - 0.362(I_{\text{Adobe 50\%}})
\end{equation}

The results of the equal slopes model can be interpreted as an increase of 0.6519 units of perceived image quality score for every additional 1 unit increase in mean training score regardless of the filter method implemented on a photo. The positive slope signifies that it is likely that the mean training score had an additive effect on image quality score given by respondents.

% add more on ANCOVA and change to APA formatting
\section{Discussion of Results}

%discussion of benchmark results
There was a lack of variation in types of methods used, which may explain the small disparities in benchmark scores between the filter methods. This could be explained by the types of algorithms implemented. These algorithms are linear filters, which are considered to be somewhat primitive in comparison to new-age neural networks or modern Transform Domain methods \citep{Mukesh}. For that reason, the $R^2$ benchmark score fluctuated no more than 0.05 units between filter methods. With the exception of SSIM, no benchmark result appeared to be lower due to image blurring. In fact, arguably the most washed photo, image $f$, received the second best marks from PSNR, MSE and $R^2$. These benchmark formulas are more likely focused on punishing random fluctuations in pixel values rather than a general skew in filtered pixel values away from their true noiseless values. 

It is not too surprising then that PSNR followed the same trends as MSE and $R^2$ in ranking the filters, especially given that its formula, equation 1.3, is not independent to either of these algorithms. However, it is interesting that MSE and $R^2$ shared trends. This is likely due to equation 1.1 and equation 1.2 relying on the residual error between the true and filtered states. SSIM had the most fascinating results. While SSIM scored the Three-by-three Mean, Bilateral, and Non-local Means filters in a similar pattern to PSNR, MSE and $R^2$, the SSIM benchmark test also scored the Adobe 50\% and Adobe 100\% filter groups shockingly low. Initial interpretation suggests that Adobe may be sacrificing pixel information to increase run time performance and structural information is being lost as a result of that drop in spatial information. Although, without more public information on the algorithm, the reason may remain unknown to the public. Studying SSIM in evaluating Adobe denoising methods surpassed the compass of this paper, but may be interesting to study in the future. Still, it is odd to see a preferred benchmark test rank an image so poorly given that the other benchmarks did not. Who is to say which is correct without more information.

%discussion of survey results
With the exception of the Three-by-three Mean Filter, all of the filter method groups in  \autoref{table:desc_survey} had image quality score means that were greater than the median. This lead to concern about a scale misunderstanding from respondents. Approximately ten percent of respondents left scores indicating that they felt the filter method had an image quality greater than the noiseless image. This result is interesting though because it signifies that respondents could not detect a significant difference between methods based on the experiment design. This could be due to the size of the image used. Perhaps a larger blown-up region or smaller original image may make individual pixel differences more apparent. Most likely, these respondents made the mistake of interpreting the maximum score as equal rather than as having superior quality to the noiseless photo. In general, it would not be very logical for one respondent to score the Adobe 100\% filter with a 2 and another to score the same filter with a score 10. The variance previously mentioned is unplanned variation within each filter group and, ultimately, lead to an inability to test for true differences between filter method image scores in the analysis of variance test. The difference in sample sizes between groups also causes the analysis of variance test to be more problematic as an assumption of that test is now violated. It would have been surprising to find sufficient evidence of population differences due to these errors in the experiment's design.

%discussion of ANCOVA
The analysis of covariance lead to an interesting finding about the unexplained variation found within image filter groups. Significantly positive slopes found in \autoref{table:un_ancova_ci} signifies that there was a linear relationship between mean training score and image quality score for the Non-local Means and Adobe 100\%. This is especially interesting because these two filters were said to be two that caused the most blurring, which can be seen in images $c$ and $f$ of \autoref{fig:images}. The only filtered training image used was a Non-local Means filter so it is possible that, in training, people who preferred blurring in image reconstitution would also rank a blurry filtered image positively later in the survey. To avoid a potential bias, future studies should randomize filter methods on training images presented to respondents as well. Regardless of the differences in slopes, there was still marginally insignificant results to justify the use of the equal slopes model. Generally speaking, there was a positive relationship between the mean training score and the final image quality score. The model shows how the Adobe filter groups are preferred based on the disparity between intercepts.

\section{Conclusion and Future Work}
In this paper, some rudimentary image denoising filters were used to process grayscale low-light photographs. When benchmark results ranked filter effectiveness differently to what the authors visually perceived, a survey was created in order to capture how others perceived the effectiveness of image denoising filters in image reconstitution. In order to compare benchmark results to survey results, descriptive statistics and an analysis of variance test were conducted. 

Descriptive statistics from \autoref{table:desc_survey} differed from the benchmark scores found in \autoref{table:benchmark}. Specifically, the benchmark tests' best ranked algorithm (the Bilateral Filter) had the lowest median score from the respondents. Additionally, there were odd results in the scoring of the two Adobe filter groups for SSIM. The analysis of variance test yielded insufficient evidence to reject the null (p=0.289), but there was reason to believe that the image quality score scale was misinterpreted by some because of the oddly high number of scores greater 5 (superior quality to the noiseless photo) collected from respondents. As a result, an analysis of covariance test was performed to quantify how the mean training score from the survey may explain some of the variation in image quality scores received from respondents in the same filter method groups. In the unequal slopes model, two significantly positive slopes were found for the mean training score's relationship with image quality scores. The Non-local Means and Adobe 100\% showed significantly positive slopes in \autoref{table:un_ancova_ci}, which gave reason to believe that the training section of the survey might have had more of an effect on the final image quality score left by respondents for these two filter methods. Marginally insignificant evidence (p=0.057) of an interaction effect between filter groups and mean training scores also warranted the use of the equal slopes model in this test. A significantly positive slope was found that quantified the mean training score's relationship with image quality scores. For every 1 unit increase in mean training score, it was expected for that the respondent score an image, regardless of filter method, 0.6519 units higher.

Future work would entail implementing modern approaches, such as a Convolutional Neural Network or a Markov Random Field, in order to produce more disparity in image reconstitution results. In addition, scale reformation should be considered so that there is less likelihood of a misinterpretation leading to bad data collection and a more informed survey population should be considered so that image quality is properly understood when rating photographs. More responses may also help discover outliers and decrease variation. Testing more than one subject matter could shed light on variation in visual perception and benchmark scores that is otherwise unnoticed. More work should be conducted in explaining why SSIM might have ranked the Adobe filters lower than even the original noisy image. Another analysis of variance test should be conducted with a dataset that has an equal number of responses per filter group. Finally, it may be interesting to complete another analysis of covariance test with the Bilateral Filter group withheld. The Bilateral Filter had the most significant difference in slope to the rest of the filter groups and was most likely the reason for the marginally insignificant p-value for interaction effect in the unequal slopes model.

\bibliographystyle{apa}
\bibliography{image_bib}

\begin{thebibliography}{}

\bibitem[\protect\astroncite{Anaya and Barbu}{2018}]{renoir}
Anaya, J. and Barbu, A. (2018).
\newblock Renoir – a dataset for real low-light image noise reduction.
\newblock {\em Journal of Visual Communication and Image Representation},
  51:144 -- 154.
\newblock
  \url{http://www.sciencedirect.com/science/article/pii/S1047320318300208}.

\bibitem[\protect\astroncite{Auger-Méthé et~al.}{2016}]{statespace16}
Auger-Méthé, M., Field, C., Albertsen, C.~M., Derocher, A.~E., Lewis, M.~A.,
  Jonsen, I.~D., and Mills~Flemming, J. (2016).
\newblock State-space models’ dirty little secrets: even simple linear
  gaussian models can have estimation problems.
\newblock {\em Scientific Reports}, 6.
\newblock \url{https://doi.org/10.1038/srep26677}.

\bibitem[\protect\astroncite{Baudes et~al.}{2005}]{baudes}
Baudes, A., Coll, B., and Morel, J. (2005).
\newblock A non-local algorithm for image denoising.
\newblock {\em 2005 IEEE Computer Society Conference on Computer Vision and
  Pattern Recognition}.
\newblock \url{https://doi.org/10.1109/CVPR.2005.38}.

\bibitem[\protect\astroncite{Bradski}{2000}]{opencv_library}
Bradski, G. (2000).
\newblock {The OpenCV Library}.
\newblock {\em Dr. Dobb's Journal of Software Tools}.

\bibitem[\protect\astroncite{Bryan and Zhao}{2017}]{googlesheets}
Bryan, J. and Zhao, J. (2017).
\newblock {\em googlesheets: Manage Google Spreadsheets from R}.
\newblock R package version 0.2.2,
  \url{https://CRAN.R-project.org/package=googlesheets}.

\bibitem[\protect\astroncite{Fernandez}{2009}]{fernandez}
Fernandez, J.-J. (2009).
\newblock Tomobflow: feature-preserving noise filtering for electron
  tomography.
\newblock {\em BMC Bioinformatics}, 10:178.
\newblock \url{https://doi.org/10.1186/1471-2105-10-178}.

\bibitem[\protect\astroncite{{Haykin}}{2001}]{939832}
{Haykin}, S. (2001).
\newblock Signal processing: where physics and mathematics meet.
\newblock {\em IEEE Signal Processing Magazine}, 18(4):6--7.
\newblock \url{https://ieeexplore.ieee.org/document/939832}.

\bibitem[\protect\astroncite{Jain et~al.}{2012}]{blur}
Jain, M., Sharma, S., and Sairam, R.~M. (2012).
\newblock Result analysis of blur and noise on image denoising based on pde.
\newblock {\em International Journal of Advanced Computer Research}, 2.
\newblock
  \url{https://pdfs.semanticscholar.org/82d5/0343f490c27fbe1a699af3d871afcc227f83.pdf?_ga=2.52751400.1426719611.1551368284-445022068.1550611831}.

\bibitem[\protect\astroncite{Kaur and Maini}{2016}]{perfeval16}
Kaur, R. and Maini, D.~R. (2016).
\newblock Performance evaluation and comparative analysis of different filters
  for noise reduction.
\newblock {\em I.J. Image, Graphics and Signal Processing}, 7:9--21.
\newblock
  \url{http://www.mecs-press.org/ijigsp/ijigsp-v8-n7/IJIGSP-V8-N7-2.pdf}.

\bibitem[\protect\astroncite{Motwani et~al.}{2004}]{Mukesh}
Motwani, M.~C., Gadiya, M.~C., Motwani, R.~C., and Jr., F. C.~H. (2004).
\newblock Survey of image denoising techniques.
\newblock {\em Proceedings of GSPx 2004}.
\newblock
  \url{https://www.cse.unr.edu/~fredh/papers/conf/034-asoidt/paper.pdf}.

\bibitem[\protect\astroncite{Plotz and Roth}{2017}]{darmstadt}
Plotz, T. and Roth, S. (2017).
\newblock Benchmarking denoising algorithms with real photographs.
\newblock {\em CoRR}, abs/1707.01313.
\newblock \url{http://arxiv.org/abs/1707.01313}.

\bibitem[\protect\astroncite{Roberts}{1963}]{larry}
Roberts, L. (1963).
\newblock Machine perception of solids.
\newblock {\em Massachusetts Institute of Technology}.
\newblock \url{http://hdl.handle.net/1721.1/11589}.

\bibitem[\protect\astroncite{{RStudio, Inc}}{2013}]{shiny}
{RStudio, Inc} (2013).
\newblock {\em Easy web applications in R.}
\newblock \url{http://www.rstudio.com/shiny/}.

\bibitem[\protect\astroncite{Russo}{2014}]{perf_eval}
Russo, F. (2014).
\newblock Performance evaluation of noise reduction filters for color images
  through normalized color difference (ncd) decomposition.
\newblock {\em ISRN Machine Vision}, 2014.
\newblock \url{http://dx.doi.org/10.1155/2014/579658}.

\bibitem[\protect\astroncite{Seletchi and Duliu}{2007}]{imgproctom06}
Seletchi, E.~D. and Duliu, O.~G. (2007).
\newblock Image processing and data analysis in computed tomography.
\newblock {\em Romanian Journal of Physics}, 52.
\newblock \url{http://www.nipne.ro/rjp/2007_52_5-6/0667_0677.pdf}.

\bibitem[\protect\astroncite{Shrivastava and Mohan}{2014}]{survey_blur}
Shrivastava, R. and Mohan, R. (2014).
\newblock Image denoising methods: A survey.
\newblock {\em International Journal of Advanced Research in Computer and
  Communication Engineering}, 3.
\newblock \url{https://www.ijraset.com/fileserve.php?FID=2818}.

\bibitem[\protect\astroncite{{van der Veen} et~al.}{2004}]{1502901}
{van der Veen}, A.~., , and {Boonstra}, A.~. (2004).
\newblock Signal processing for radio astronomical arrays.
\newblock In {\em Processing Workshop Proceedings, 2004 Sensor Array and
  Multichannel Signal}, pages 1--10.
\newblock \url{https://ieeexplore.ieee.org/document/1502901}.

\bibitem[\protect\astroncite{Zhang}{2015}]{zhang}
Zhang, Z. (2015).
\newblock Image noise: Detection measurement and removal techniques.
\newblock
  \url{https://pdfs.semanticscholar.org/89b8/55e8373674d0ca47be25eb82358710fee15d.pdf?_ga=2.215975959.1426719611.1551368284-445022068.1550611831}.

\end{thebibliography}
\end{document}